\pdfoutput=1

\documentclass[11pt]{article}

\usepackage{ACL2023}

\usepackage{times}
\usepackage{latexsym}
\usepackage[T1]{fontenc}

\usepackage[utf8]{inputenc}

\usepackage{microtype}
\usepackage{float}
\usepackage{booktabs}
 \usepackage{graphicx}
 \usepackage{amsmath}
 \usepackage{enumitem}
 \usepackage{arydshln}
 \usepackage{amssymb}
 \usepackage{subfig}
 
\usepackage{inconsolata}

\usepackage{csquotes}

\newcommand{\de}{\texttt{DE\textsubscript{train}}}
\newcommand{\dach}{\texttt{DACH\textsubscript{train}}}

\setlist[itemize]{noitemsep, topsep=0pt}

\title{Additive manifesto decomposition: \\ A policy domain aware method for understanding party positioning}

\author{First Author \\
  Affiliation / Address line 1 \\
  Affiliation / Address line 2 \\
  Affiliation / Address line 3 \\
  \texttt{email@domain} \\\And
  Second Author \\
  Affiliation / Address line 1 \\
  Affiliation / Address line 2 \\
  Affiliation / Address line 3 \\
  \texttt{email@domain} \\}

\author{Tanise Ceron \qquad Dmitry Nikolaev \qquad Sebastian Pad\'o \\
  Institute for Natural Language Processing, University of Stuttgart, Germany  \\
  \texttt{\{tanise.ceron,dmitry.nikolaev,sebastian.pado\}@ims.uni-stuttgart.de} \\
}

\begin{document}

\maketitle

\begin{abstract}

Automatic extraction of party (dis)similarities from texts such as party election manifestos or parliamentary speeches plays an increasing role in computational political science.
However, existing approaches are fundamentally limited to targeting only \textit{global} party (dis)\-similarity: they condense the relationship between a pair of parties into a single figure, their similarity. In aggregating over all \textit{policy domains} (e.g., health or foreign policy), they do not provide any qualitative insights into which domains parties agree or disagree on.

This paper proposes a workflow for estimating policy domain aware party similarity that overcomes this limitation. The workflow covers (a)~definition of suitable policy domains; (b)~automatic labeling of domains, if no manual labels are available; (c)~computation of domain-level similarities and aggregation at a global level; (d)~extraction of interpretable party positions on major policy axes via multidimensional scaling. 
%
We evaluate our workflow on manifestos from the German federal elections. We find that our method (a)~yields high correlation when predicting party similarity at a global level and (b)~provides accurate party-specific positions, even with automatically labelled policy domains.





\end{abstract}

\section{Introduction}
\label{sec:intro}
Party competition is a fundamental process in democracies. It provides space for different political stances to emerge, allowing people to choose which of them they most identify with. Investigating this process is relevant for understanding the reasons behind the choice of voters in elections as well as the behavior of parties in policy decision-making once in power \citep{benoit2006party}. 

Within political science, the positioning of parties is investigated under the umbrella term of \enquote{party competition}. Some studies look at specific policies such as \enquote{welcoming refugees}, others, at broader domains such as \enquote{economy}. Traditionally, the positioning of parties within these policies or domains is scaled down to a reduced number of political dimensions such as the well-established left-right or the libertarian-authoritarian axes in order to facilitate the comparison among parties and their ideologies \citep{heywood2021political}. Analyses are usually carried out by experts, who gather policy and ideological stances of members of the political parties in several countries in Europe and beyond \citep{jolly2022chapel}. Alternatively, electoral programs are manually annotated following a specific codebook that takes into account the position of the parties on policies so that the salience of the labels can be scrutinised \citep{manifesto-burst}. 

Recently, computational approaches have 
been developed to automate and scale up party position analysis to larger amounts of text \citep{slapin2008scaling, daubler2021scaling, ceron-2022}. This development has the potential of alleviating the burden of annotation, but has so far been realised only at 
an \textit{aggregated} level: party positions are projected on the left-right scale or on a distance-based approach between party pairs according to several policies, not providing insights at the level of policy domains. This requires political scientists either to manually check for sections of the text of their interest in case the objective is to understand the positioning of parties on a more fine-grained level or to make assumptions about a policy considering the entire document. 

\begin{table*}[th!]
\centering
\setlength\tabcolsep{4pt}
\begin{tabular}{lll}
\hline
\textbf{Party} & \textbf{Text}                                                                                                                                                 & \textbf{Category}                                                        \\ \hline
AfD            & The principles of equality before the law.                                                                                                                    & Equality: Positive                                                       \\
CDU            & We are explicitly committed to NATO's  2\% target.                                                                                                            & Military: Positive                                                       \\
FDP            & \begin{tabular}[c]{@{}l@{}}And with a state that is strong because it acts lean and modern \\ instead of complacent, old-fashioned and sluggish.\end{tabular} & \begin{tabular}[c]{@{}l@{}}Government and \\ Admin. Efficiency\end{tabular} \\
SPD            & \begin{tabular}[c]{@{}l@{}}There need to be alternatives to the big platforms - with real \\ opportunities for local suppliers.\end{tabular}                  & Market Regulation                                                        \\
Grüne          & We will ensure that storage and shipments are strictly monitored.                                                                                             & Law and Order: Positive                                                  \\
Die Linke      & Blocking periods and sanctions are abolished without exception.                                                                                               & Labour groups: Positive                                                  \\ \hline
\end{tabular}
\caption{Translated examples of sentences from German federal election manifestos (2021)  with their categories as annotated by the Comparative Manifesto Project.}
\label{tab:manifestos}
\end{table*}

In this paper, we extend the previous studies to provide a computational model for party positions and party similarity \textit{at the level of policy domains}. To do so, we semi-automatically decompose the texts into interpretable thematic blocks based on an updated inventory of annotated labels from the Comparative Manifesto Project (CMP). Sentence embeddings leverage well the grouping of finer-grained categories into these blocks, which we call policy domain from now on. Then, they are used to compute pairwise policy differences between parties. The results show that this re-grouping of categories into higher policy domains performs well not only at an aggregate level in comparison with the ground truths, but that they also match the positioning of parties within the political dimensions at the individual level of policy domains.

Besides shedding light on the positioning of parties regarding where they most (dis)agree, we also avoid  relying on the \textit{salience} (i.e., frequency) of the categories. This assumption is implicit in many existing party positioning models including our own prior work \citep{ceron-2022} and is motivated on the grounds that major domains, such as economic and social policy, should play a more prominent role. At the same time, there is strong evidence that voters re-weigh domains by their priorities \citep{10.2307/2111335}. We take this as evidence that models would benefit from focusing on modeling \textit{within}-domain similarities and differences between parties.

We evaluate the extent to which annotations can be forgone by evaluating several classifiers to automatically predict the policy domains of the 2021 German federal elections 
based on annotated manifestos from previous elections. Comparing the party positioning given by the manually annotated and the predicted labels, we find that the classifier can substitute annotations at an aggregate level and also in most policy domains, allowing new, unannotated documents to be analysed automatically. We make our code freely available.\footnote{ \url{https://github.com/tceron/additive_manifesto_decomposition} }


\section{Related Work}
\label{sec:related-work}

\paragraph{The Comparative Manifesto Project.}
Party manifestos, also known as electoral programs or party platforms,
condense parties' ideologies and stances towards various policies
\citep{budge2003validating}.  The Comparative Manifesto
Project\footnote{\url{https://manifesto-project.wzb.eu/}} annotates
manifestos from multiple countries around the world following a
codebook that takes into account the positioning of parties according
to the left-right political dimension \citep{budge2001-mapping}. The
codebook contains 143 fine-grained categories. Table
\ref{tab:manifestos} shows some examples. The categories are labelled
according to policies and may or may not contain the stance towards
the policy as well. For example, there are two labels for
\textit{Military}: \textit{Military: Positive} and \textit{Military:
Negative}, but there is only one category for \textit{Peace} because
no party is against it. In most cases, the annotations are assigned to
every sentence of the manifesto, however, sentences are split into
smaller parts whenever there is more than one self-contained category.

\paragraph{Computational models of party positioning.}
Party manifestos, which provide a particularly rich source of
information on parties' positions, have been extensively used in computational political science. In the pre-neural era, they mainly focused on word/token distributions to position parties along a scale; thus, the Wordscore approach used the distributions extracted from reference texts to determine party positioning of new texts \citep{laver2003extracting}. \citet{slapin2008scaling} focus on overcoming the disadvantageous dependence on reference texts which  assumes that political discourse does not change significantly over time and that the reference corpus always contains  good representations of extreme policy positions.

Arguably, the adoption of (static) word embeddings such as Word2Vec \citep{Mikolov2013EfficientEO} instead of word distributions constituted a step forward for computational models of party positioning. For example, \citet{glavas-etal-2017-unsupervised} 
take advantage of the possibility to align word embeddings across languages to present a multilingual model for extracting party positions from speeches of the European parliament. \citet{rheault2020word} exploit
another property of embedding spaces, namely the information on graded word similarity implicit in them. 
They build combined representations from word embeddings and political metadata and then estimate the positions of different parties through dimensionality reduction. The embeddings are reduced to two dimensions and their projection in the space shows the alignment of parties from Britain, Canada, and the US on a left-right axis.

The recent shift from static word embeddings to contextualized embeddings was a second important step. Contextualized embedding models, like BERT \citep{devlin-etal-2019-bert}, are not only able to pick up on corpus-specific usage of words, but can also be fine-tuned for specific tasks, which greatly improves the quality of the representations. In previous work \citep{ceron-2022}, we predicted global party similarity using Sentence-BERT (SBERT, \citealt{reimers-gurevych-2019-sentence}), a model for the task of sentence-similarity prediction. It uses a Siamese network with a triplet loss function that aims at placing mutually similar sentences close to one another in embedding space and pushing dissimilar ones apart. We found that SBERT representations can profit substantially from tuning by party, forcing the model to place sentences from the same party closely together in the semantic space. 

 Architectures similar to SBERT with  modifications in the loss function have followed such as different types of contrastive and non-contrastive self-supervised learning \citep{gao-etal-2021-simcse} and normalization techniques in the distribution through an unsupervised objective during training \citep{li-etal-2020-sentence}. The original SBERT architecture, though, remains the most widely used and numerous pre-trained models, including multilingual ones, have been made publicly available \citep{ceron-2022}.
 

Despite these successes, the computational studies mentioned above have not proposed a general way of capturing the positioning of parties within specific policy domains, 
opting for narrowly applicable ad-hoc modifications of existing algorithms. For example, \citet{laver2003extracting} adapt their reference values (related to the word distribution) to few chosen domains, and \citet{slapin2008scaling} manually identify sections of the manifestos that discuss economic issues.

\section{Methodology}

\subsection{Workflow}

\begin{figure}[t]
    \centering
    \includegraphics[width=0.5\textwidth]{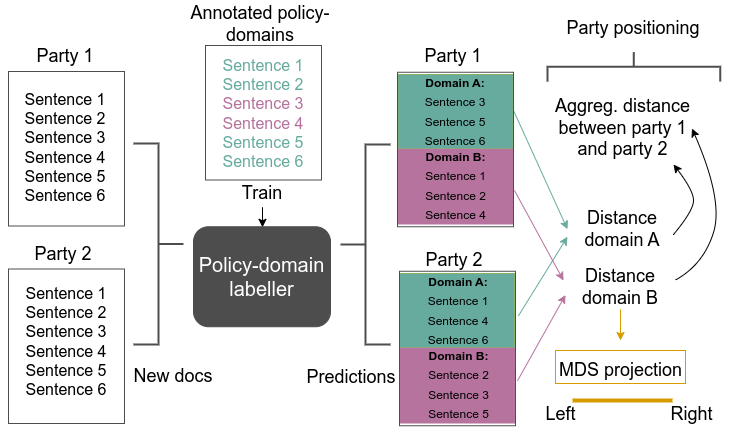}
    \caption{The workflow of additive manifesto decomposition for party positioning analysis.}
    \label{fig:workflow}
\end{figure}

The goal of the additive manifesto decomposition method we propose is to computationally analyse the positioning of parties both at the level of policy domains and at an aggregated level of information. Figure \ref{fig:workflow} illustrates the four steps in which we decompose this analysis: (1), we define policy domains (visualized as colors). This is discussed in Section \ref{sec:policydomaingrouping}. (2), we label
 manifestos with the policy domains. Unless manual annotation is available, this involves training a policy domain labeller. This is discussed in Section \ref{sec:policydomainprediction}. (3),  we represent parties' positions on policy domains by vectors
 and  compute the similarities between these vectors,  which can later be aggregated to obtain global similarities. This is discussed in Section \ref{sec:computingparty}.
Finally, (4), we apply a dimensionality reduction technique to the parties' policy domain distance matrix to be able to inspect their positions.  

We apply the methods that we propose to corpora from the Comparative Manifesto Project (CMP, cf. Section~\ref{sec:related-work}) and use examples from the CMP below for illustration. However, we believe that the CMP is fairly typical regarding size and annotation granularity for  resources in computational political science. We are confident that our methods generalize to other corpora.

\subsection{Policy Domain Grouping}
\label{sec:policydomaingrouping}

Given that the objective is to understand where parties (dis)agree the most according to the way they expose their stances and ideologies in the manifestos rather than on the salience of mentions of a policy, we first have to decompose the manifestos into interpretable thematic blocks, which we identify as policy domains.
Policy domains are in principle freely definable in an inductive fashion \citep{Waldherr_Wehden_Stoltenberg_Miltner_Ostner_Pfetsch_2019}
but must fulfil three  requirements to be useful:
\begin{itemize}
    \item[(1)] Domains must be coherent and interpretable in the context of policies to support the goal of understanding in which domains parties are most similar and dissimilar
    \item[(2)] Domains must be neutral with regard to stance. In other words, the categories with opposite stances (positive and negative) vis-a-vis a certain problem (e.g., immigration) should belong to the same policy domain.
    \item[(3)] Domains must be located at the right level of granularity: they must be detailed enough to be informative (cf. (1)), but not so detailed that accurate classification becomes impossible in practice. For example, the original CMP categories are arguably too fine-grained (such as the examples in Table \ref{tab:manifestos}).
\end{itemize}
%
%
We propose that a reasonable granularity for party positioning 
can typically be achieved by \textit{clustering} fine-grained
category annotations from sources such as the CMP codebook. 

To do so, we represent the texts through sentence embeddings as state-of-the-art representations (cf. Section \ref{sec:related-work}). This already enables us to compute cosine distances between all pairs of sentences belonging to two categories and use their average as a distance measure of topical coherence between two given categories. Formally, given a set of sentences $\{s_1, s_2, \dots, s_n\}$ and a disjoint collection of categories $\{C_1, C_2, \dots, C_k\}$, for each category pair $(C_p, C_q)$, we compute $$\operatorname{dist}(C_p, C_q) = \dfrac{1}{N} \sum_{i \in C_p, j \in C_q} 1 - \operatorname{cosine}(s_i, s_j)$$ where $N$ is the number of sentence pairs.

The resulting distance matrix between low-level CMP categories can then serve as input for an average-linkage hierarchical-clustering algorithm, which produces a tree of categories, from which a suitable level of abstraction can be selected that meets the requirements laid out above. Inspection of candidate policy domains is also adopted as a sanity check for the sentence embedding model.

\newcommand{\psim}{\mathop{\mathrm{pair_s_i_m}}}
\newcommand{\catsim}{\mathop{\mathrm{cat_s_i_m}}}

\subsection{Policy Domain Prediction}
\label{sec:policydomainprediction}

For texts without policy domain annotation, we predict policy domains for all sentences using existing annotated corpora as training data. Technically, this is a labeling task where each token is a sentence (or segment thereof) which can be solved by any state-of-the-art classifier architecture. It has two main challenges. The first one is  the high contextual dependence on political discourse. As a result, the classification of individual sentences is often challenging. For example, a vague formulation, such as \textit{There is still a lot to do}, must take into account based on the category of the previous sentence, a possibility explicitly acknowledged by the CMP codebook. This clearly indicates that it is sensible to approach domain prediction as a \textit{sequence} labeling task.

 The second challenge is that training and test data are always bound to be \enquote{out of domain}, since they will differ in either country or time: we either need to project from past elections to new ones, or across countries, and thus political cultures. Since both of these settings can introduce strong concept drift, this makes the task an example of out-of-domain prediction.

The end result of policy domain prediction is then a decomposition
of a party manifesto $p$ into a disjoint collection of $k$ policy domains $\{D^p_1, D^p_2, \dots, D^p_k\}$. Note that the set of sentences associated with any domain may be empty.

\subsection{Computing Party (Dis)similarities}
\label{sec:computingparty}
After decomposing the sentences of manifestos into policy domains, we  compute the similarity between parties by domain. We re-use the simple coherence measure from
the policy domain grouping (cf. Section~\ref{sec:policydomaingrouping}). Again, this involves choosing a sentence embedding model, a parameter of our method. Given two parties' manifestos $p$ and $q$, we interpret $dist(D^p_i,D^q_i)$, the average pairwise distance among sentences for policy $i$ as the 
 distance between parties $p$ and $q$ for this domain.

%

To obtain an aggregated party distance, we simply \textit{average} the distances of all policy domains.
As argued in Section~\ref{sec:intro},
this removes the effect of domain salience from the model and arguably obtains the clearest party positioning as perceived by a \enquote{neutral} voter \citep{10.2307/2111335}.

\subsection{Multidimensional Scaling}
\label{sec:mds}

The results of the previous step can be represented as a square matrix of the distances between every party pair.
In order to enable a more qualitative analysis of the results by policy domain, we apply a multi-dimensional scaling (MDS) technique which maps a distance matrix onto a one-dimensional scale while respecting the distances as well as possible. MDS models are well established for visualization in political science \citep{rheault2020word,heywood2021political}. We utilize  Principal Component Analysis 
is chosen because the first component explains well the variability in the data. 


\section{Experimental Setup}

\subsection{Data}

We analyze the positions of the six German parties which obtained parliamentary seats in 2021 based on their 2021 federal election manifestos. These are Die Linke, Bündnis 90/Die Grünen, Christian Democratic Union (CDU), Free Democratic Party (FDP), Social Democratic Party for Germany (SPD), and Alternative for Germany (AfD).

We train a policy domain labelled for these manifestos based on the annotated data provided by the CMP. We experiment with two training sets: \de\ contains only manifestos from Germany dating from 2002 to 2017. The second instead,  \dach\ consists of manifestos from the majority German-speaking countries (Germany, Austria, and Switzerland) for all elections from 2002 to 2019. This allows us to understand whether the classifier benefits more from focused data of a single country (the country of interest for the analysis) or if the raw amount of data is more 
relevant. Appendix~\ref{sec:appendix-a} provides details on data statistics.

\subsection{Policy Domain Grouping}
\label{subsec:policy_grouping}

To define our policy domains, we concatenate the manifestos of six German major political parties from the 2021 elections, together with their CMP annotations, into a single corpus. It contains a total of 69 annotated categories, however, only the ones with 10 occurrences or more are included in the grouping - a total of 61. We employ \texttt{multilingual-mpnet-base-v2}, the vanilla SBERT model to compute similarities\footnote{Provided by HuggingFace as a part of \texttt{sentence-transformers} collection.} in order to make the clustering more general.  It is a vanilla multi-lingual model with the base-size version of XLM-RoBERTa \citep{conneau-etal-2020-unsupervised} as the encoder trained on more than 50 languages.\footnote{\url{https://huggingface.co/sentence-transformers/paraphrase-multilingual-mpnet-base-v2}}

Representations from the multilingual SBERT model are post-processed with whitening
transformation \citep{Su2021WhiteningSR}, as suggested by
experiments finding that more isotropic embeddings capture political
text similarity substantially better \citep{ceron-2022}.

Hierarchical agglomerative clustering led to a clustering that consistently grouped thematically close categories with opposite valences into single domains, as shown in Fig.~\ref{fig:hirarchical-clustering} in Appendix \ref{sec:appendix-b}. In the inspection of the clustering tree, we verify that all 10 categories that contained positive and negative labels fall in the same cluster in order to satisfy requirement 2. We then selected the tightest possible clusters of categories that together formed coherent policy domains (fulfilling requirements 1 and 3). The remaining 8 categories (that were not included in the clustering) are added to the formed clusters manually. We consulted with political scientists and related work \citep{benoit2006party, jolly2022chapel} to verify the result. The full list of CMP categories falling into each of our issues is presented in Appendix \ref{sec:appendix-b}.

\subsection{Policy Domain Labelling}


As stated above in Section\ \ref{sec:policydomainprediction}, domain labels
in manifestos are context-dependent. Therefore, we
give up the assumption of previous analyses of manifestos \citep{daubler2021scaling} that annotated sentences are independent units of information.
Instead, we treat policy domain labelling as a sequence labelling task. Our preliminary experiments showed that incorporating sequence information is indeed beneficial for prediction quality, and we chose a simple \enquote{bigram}-based model: pairs of subsequent sentences from manifestos were concatenated, and the model was tasked with predicting the label of the second one.\footnote{I.e.\ we are not using the label of the first sentence. Using it could help with training but may lead to increased variance on new data where an incorrect label for a sentence would then bias the prediction for the next one.}

We use averaged token embeddings from \texttt{xlm-roberta-large} and pooled representations from the multilingual version of \texttt{mpnet-base-v2} fine-tuned on paraphrase detection as sentence-pair embeddings\footnote{\texttt{xlm-roberta-large} is nearly twice as big as the sentence transformer but benefited from less sentence-focused training.} as encoded representations and use a two-layer MLP with tanh activation as the classification head. 
The system is then trained end-to-end for two epochs.
As a first baseline, we choose the majority baseline between the 14 categories (13 policy domains in addition to the category \enquote{Other} which does not belong to any domain). The second baseline instead follows the same bi-gram idea in terms of input and is logistic regression fed with the representation taken from frozen SBERT \texttt{mpnet-base-v2}. 

\subsection{Party (dis)similarity -- sentence encoders}

 We experiment with four different sentence encoding models when computing party similarities (as explained in Section \ref{sec:computingparty}). Our baseline is FastText for German based on character n-gram embeddings \citep{bojanowski2017enriching}.\footnote{Pre-trained model downloaded from \url{fasttext.cc}} The second model is a base-sized cased version of BERT trained on German data, a monolingual Transformer-based model. The representation of a given sentence from these models is an average of its token embeddings. Then, as end-to-end sentence encoders we use two versions of SBERT. The first is the vanilla SBERT pre-trained model \texttt{multilingual-mpnet-base-v2}. The second is SBERT$_{\text{domain}}$, a pre-trained model from our prior work \citep{ceron-2022}, which we fine-tuned on German CMP data from before 2019 to distinguish between 6 higher-level domains from the CMP codebook. 

 Our preliminary experiments showed that applying post-processing with whitening improves all models. Therefore, all sentence representations in this step are whitened as in Section \ref{subsec:policy_grouping}.


\subsection{Evaluation}

\subsubsection{Ground Truth}

We evaluate our additive manifesto decomposition method against two sources of ground truth. 

\paragraph{RILE index.} The RILE index is a widely used way of computing the positioning of parties on certain policy domains or in aggregate. \citet{laver1992measuring} selected 12 categories from the CMP codebook as left-leaning and 12 others as right-learning.\footnote{The table of categories can be found at \url{https://manifesto-project.wzb.eu/down/tutorials/main-dataset.html}} The score is then computed as $\operatorname{RILE} = (R - L) / N$, where $R$ and $L$ are counts of sentences from the right and left categories, respectively. Dividing by $N$, the manifesto length, results in a normalized score between -1 and 1.

As our approach returns a distance matrix, we need to use dimensionality reduction to obtain a single estimate per party. For this purpose, we extract the first axis of the classical MDS algorithm applied to distance matrices -- corresponding to the first principal component in PCA analysis. 

\paragraph{CMP-category salience.} Given that the RILE index makes use of only 24 out of the 143 categories from the CMP codebook, we used another type of ground truth that takes into account all categories and corresponds to the traditional political science approach of comparing domain saliences, i.e.\ relative prominences of different policy categories in manifestos \citep{budge2001-mapping}. Each party is represented as a vector of relative frequencies of categories normalized by the manifesto length. Euclidean distances between these representations are then used to create a party distance matrix. 

\subsection{Evaluation Metrics}

We evaluate the results of the first principal component analysis against the RILE score with Pearson correlation in order to understand the extent to which our models capture the aggregated left-right dimension of the political spectrum through textual similarity. For checking how well our method captures the more nuanced method of measuring party-platform dissimilarities from category saliences, we use the Mantel test \citep{mantel1967detection}. For both metrics, both by-domain and aggregate agreement scores can be computed.

For experiments with unannotated manifestos, we predict the policy domain labels using the best-performing classifier and then repeat the evaluation in the same way using the predicted labels.



\section{Results and Discussion}

\subsection{Annotated Setup}

In the \textit{annotated setup}, we use the ground truth of policy
domains as annotated in the CMP dataset. We evaluate
party-positioning landscape extracted using our method, both in
aggregate and for different policy domains, against the ground truths:
the RILE scores and the distances computed using CMP-category
saliences.

\paragraph{Aggregated similarity.} Table \ref{tab:results-aggreg} illustrates the correlation of the aggregated similarity computation with the ground truths. Correlations are very high in both ground truths with small differences across models. FastText, our baseline, performs best in predicting the Rile index (Mantel $r$ = 0.94) and second in the CMP distance ($r$ = 0.80). We believe that the excellent performance of this model is given due to the similarity computation. The comparison between sentences from the same policy domain (theme) might help in capturing fine-grained differences in stances between parties. BERT$_{\text{German}}$ is the model that performs the worst even though for a slim difference -- as previous research suggested, the quality of BERT for sentence representation is low \cite{li-etal-2020-sentence}. Finally, SBERT$_{\text{vanilla}}$ and SBERT$_{\text{domain}}$ have comparable results. While the former performed the best on RILE ($r$ = 0.91) in comparison with the latter ($r$ = 0.87), the latter comes out first in the CMP distances ($r$ = 0.84 vs.\ 0.80). This suggests that the non-fine-tuned model can still excel in the task of text similarity on out-of-domain data. Depending on the purpose, however, the fine-tuned version might be a better option, in line with previous results on representing political text \citep{ceron-2022}. 


\begin{table}[t]
\centering
\setlength\tabcolsep{5pt}
\begin{tabular}{lllll}
        & \multicolumn{4}{c}{\textbf{Policy Domains are \dots}} \\
                  & \multicolumn{2}{c}{\textbf{Annotated}}                                                                                                                          & \multicolumn{2}{c}{\textbf{Predicted}}                                                                                                                          \\ \hline
\textbf{Model}    & \multicolumn{1}{c}{\textbf{\begin{tabular}[c]{@{}c@{}}Rile\\ ($r$)\end{tabular}}} & \multicolumn{1}{c}{\textbf{\begin{tabular}[c]{@{}c@{}}CMP\\ (Man.)\end{tabular}}} & \multicolumn{1}{c}{\textbf{\begin{tabular}[c]{@{}c@{}}Rile\\ ($r$)\end{tabular}}} & \multicolumn{1}{c}{\textbf{\begin{tabular}[c]{@{}c@{}}CMP\\ (Man.)\end{tabular}}} \\ \hline
FastText          & \textbf{0.94*}                                                                    & 0.80*                                                                             & 0.67                                                                              & 0.76*                                                                             \\
BERT$_{\text{German}}$   & 0.84*                                                                             & 0.77*                                                                             & 0.59                                                                              & 0.79*                                                                             \\
SBERT$_{\text{vanilla}}$ & 0.91*                                                                             & 0.80*                                                                             & 0.56                                                                              & 0.71*                                                                             \\
SBERT$_{\text{domain}}$  & 0.87*                                                                             & \textbf{0.84*}                                                                    & \textbf{0.79*}                                                                    & \textbf{0.80*}                                                                    \\ \hline
\end{tabular}
\caption{Correlations of party distances produced by our method with ground truths. For comparison with the RILE index, the first axis of an MDS projection computed based on the distance matrix is used. CMP domain-based distances form their own distance matrix. *~means \textit{p} < 0.05.}
\label{tab:results-aggreg}
\end{table}

\begin{figure}[tb!]
  \centering
  \subfloat{\includegraphics[width=\columnwidth]{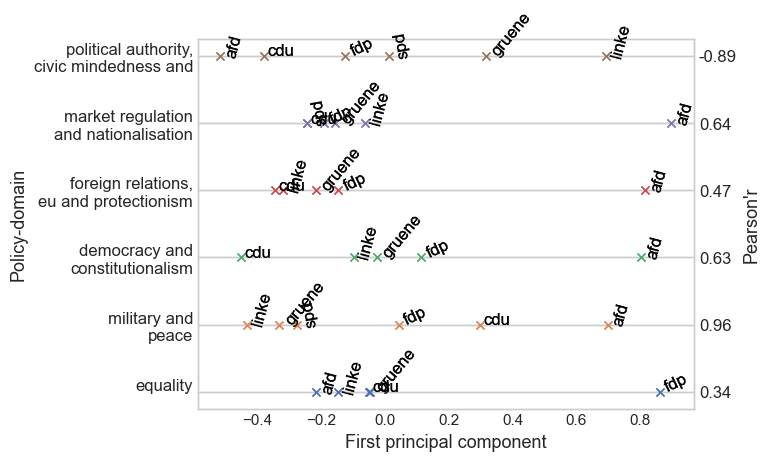}\label{fig:f1}}

\subfloat{\includegraphics[width=\columnwidth]{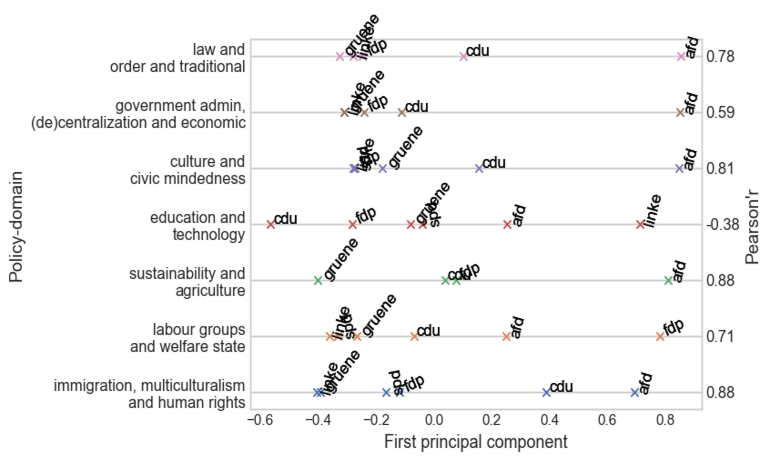}\label{fig:f2}}
  \caption{First axis of MDS projections derived from the SBERT$_{\text{domain}}$ by-policy-domain distance matrices. Pearson $r$ values give correlation to Rile scores. See Appendix \ref{sec:appendix-positioning} for  full party names.}
  \label{fig:mds}
\end{figure}

\paragraph{Similarity by policy domains.}
We further analyze the output of the best model, namely SBERT$_{\text{domain}}$. 
Figure~\ref{fig:mds} shows the results of the application of MDS to the policy domain distance matrices. On the left-handed side of the plot lies the name of policy domains and on the right-handed side the Pearson's $r$ with respect to the RILE score. 

The higher the (absolute value of the) correlation coefficient, the
more the scale in question follows the classic left-right scale as
measured by RILE.  As expected, some policy domains yield high
correlation whereas others do not. Importantly, this is not a measure
of model quality. Rather, as it has often been observed in the
political-science literature, the left-right scale cannot explain the
complete picture of party positioning
\citep{heywood2021political}. Therefore, quantitative analysis has to
be complemented by qualitative judgments about the appropriateness of
the predictions.

Indeed, the results mirror some well-known facts about German
politics. For example, in \textit{foreign relations, EU and
protectionism} -- which is only moderately correlated with the
left-right scale at $r$ = 0.47 -- the AfD is an outlier compared to
other parties, arguably because it is against being part of the
European Union and has a different stance with regard to having ties
with Russia as compared with the other parties, which all fall in the
same region. Another case is \textit{education and technology} where
AfD and Die Linke, who are generally can be regarded as the opposite
pole of the left-right spectrum, happen to share a lot of common
ground in their stance toward the expansion of education and investment in
technology and infrastructure ($r$ = -0.38). On the other hand, in
policy domains such as \textit{military and peace} and
\textit{immigration and multiculturalism}, party positions align very
well with the overall left-right scale ($r >$ 0.85), with right-leaning
parties being more militaristic and immigration averse.

In sum, we take the results of this analysis as evidence that our workflow produces accurate fine-grained characterizations of party positions.

\subsection{Predicted Setup}

In the \textit{predicted setup}, we do not use the CMP annotations of
policy domains but predict the policy domains instead.

\paragraph{Policy domain labeller.}

Table \ref{tab:acc-classifier} shows the accuracy of the models and the majority baseline on the test set. Overall, the larger but more varied training set including all German-speaking countries (\dach) performs better than \de\ (data only from Germany) in all models, suggesting that it is not necessary to exclusively have data from the same country of analysis -- given the similarity in the political scenario. As expected, the SBERT$_{\text{frozen}}$ which is not fine-tuned for the task, performed the worst (55.3\% and 56.7\%). Whereas SBERT+MLP performed second (60.4\% and 63.1\%) and the best model is XLM-RoBERTa-large+MLP (62.5\% and 64.5\%), whose bigger size likely won over additional pretraining of a smaller model.
The results of the XLM-RoBERTa-large model fine-tuned on \dach\ are used for the rest of this analysis. 

\paragraph{Aggregated similarity.}
We evaluate how the predictions of our policy domain labeller perform in a scenario where there are new upcoming elections and no annotations are available. Table~\ref{tab:results-aggreg} shows that even though even results are not as incisive as in the annotated scenario, the correlation scores are still high for CMP saliences. In terms of models, SBERT$_{\text{domain}}$ is the best-performing model (Mantel $r$~=~0.80), similarly to the annotated scenario
SBERT$_{\text{vanilla}}$ is the worst performing encoder ($r$~=~0.71), with  FastText ($r$~=~0.76) and BERT$_{\text{German}}$ ($r$~=~0.79) in between. As for the RILE score, only SBERT$_{\text{domain}}$ demonstrates a statistically significant correlation. These results confirm that the additive manifesto decomposition is dependent on the precision of the policy domains labels but can also provide interpretable results for unannotated data.

\begin{table}[tb]
\centering
\begin{tabular}{lcc}
\hline
\textbf{Model}           &  \textbf{\de\ } & \textbf{\dach\ } \\ \hline
Majority Baseline        & 14.5\%           & 14.5\%                   \\
SBERT$_{\text{frozen}}$+log. reg. & 55.3\%                 & 56.7\%                   \\
RoBERTa$_{\text{xlm}}$+MLP          & \textbf{62.5\%}           & \textbf{64.5\%}                   \\
SBERT$_{\text{tune}}$+MLP      & 60.4\%           & 63.1\%                   \\ \hline
\end{tabular}
\caption{Accuracy score of the classifier on the test set (same test set for both training datasets). }
\label{tab:acc-classifier}
\end{table}

\begin{table}[t]
\centering
\begin{tabular}{p{4cm}ll}
\textbf{Policy domain}                                                                       & \multicolumn{1}{c}{\textbf{Mantel}} & \textbf{Acc.} \\ \hline
culture and civic mindedness                                                                 & 0.51                                & 58.2\%        \\ \hline
democracy and constitutionalism                                                              & 0.92*                               & 62.8\%        \\ \hline
education and technology                                                                     & 0.89*                               & 61.8\%        \\ \hline
equality                                                                                     & 0.94*                               & 70.7\%        \\ \hline
foreign relations, eu and  protectionism           & 0.96*                               & 70.5\%        \\ \hline
government admin, (de) centralization and econo... & 0.91*                               & 53.0\%        \\ \hline
immigration, multiculturalism and human rights    & 0.96*                               & 53.8\%        \\ \hline
labour groups and  welfare state                   & 0.69*                               & 72.7\%        \\ \hline
\begin{tabular}[c]{@{}l@{}}law and order and \\ traditional morality\end{tabular}            & 0.78*                               & 71.8\%        \\ \hline
\begin{tabular}[c]{@{}l@{}}market regulation and \\ nationalisation\end{tabular}             & 0.83*                               & 72.0\%        \\ \hline
military and peace                                                                           & 0.88*                               & 86.9\%        \\ \hline
\begin{tabular}[c]{@{}l@{}}political authority, \\ civic mindedness and anti...\end{tabular} & 0.34                                & 27.9\%        \\ \hline
\begin{tabular}[c]{@{}l@{}}sustainability and \\ agriculture\end{tabular}                    & 0.97*                               & 77.4\%        \\ 
\end{tabular}
\caption{Mantel correlation between the distance matrices of the annotated and the predicted setups. $*$ means $p <$ 0.05. Acc.: accuracy of classifier within each policy domain. }
\label{tab:by-policy-mantel}
\end{table}

\paragraph{Similarity by policy domains.}

Our sources of ground truth do not provide us with gold measures of the similarity within each policy domain. Therefore, we cannot directly evaluate by-domain matrices produced with the predicted data. However, we can indirectly evaluate their usefulness by comparing them to the matrices produced using the gold annotations, which we already know to be highly meaningful.

Table \ref{tab:by-policy-mantel} shows the Mantel correlations between the
distance matrices produced with the annotated setup and the
one from the predicted setup for each policy domain. Mantel
correlation is 0.78 or higher in 10 out of 13 policy domains. Negative
outliers are \textit{culture and civic mindedness}, \textit{political
authority} and \textit{labour groups and welfare state}.
We further investigate whether there is a correlation between the
number of correctly labelled sentences by classifier (measured by
accuracy) and Mantel correlation of the results. We find that there is
a relatively strong correlation (Pearson $r$~=~0.59, $p$~=~0.03). This
suggests that one can predict which policy domains will yield the most
faithful results in an unsupervised scenario on the basis of their
accuracy in the policy domain labeling part of the workflow.

\section{Conclusion}
\label{sec:conclusion}

In our first contribution, we introduce Additive Manifesto Decomposition, 
a workflow for efficient analysis of party platforms, both in aggregate and across a range of policy issues. It builds on state-of-the-art sentence-representation models, which it uses for three operations on policy domains: definition, prediction, and (cross-party) similarity computation. In this manner, our workflow can incorporate advances on the representational level \citep{reimers-gurevych-2019-sentence,ceron-2022} but complements them with a crucial level of reflection and analysis at the informative level of policy domains.

Our second contribution is a study of the political
landscape in Germany using our workflow. The results we obtain match well with expert judgements,  suggesting that our workflow  yields a reliable technique to automatically study the similarity between parties across policy domains. In addition to analysing the implicit stance space, operationalized through distance matrices derived from text similarity, we show that our method makes it possible to recover the traditional scaling analyses of the political science literature: we can efficiently approximate the aggregate RILE (right-left) scores provided by experts in the aggregate settings, and when proceeding by domain, we see that our methods recover non-trivial policy configurations, e.g., the agreement of the far-right and far-left parties in Germany on the subject of EU and the expansion of education. Moreover, we show that classifiers substitute the annotations of these high-level domains and still yield similar results as compared to the fully annotated scenario. 

 Germany provided an appropriate target for our case study, given both the large number of annotated manifestos and large body of expert analyses. Nevertheless, an important direction for future work is testing the applicability of our workflow to other countries, in particular regarding the training of policy domain labelers given the challenging concept drift between elections, and the possible cross-lingual application of our model components despite differences between political cultures \citep{doi:10.1177/1354068818805248}.

 Lastly, our methodology does not only suit the identification of the positioning in the political domain, but more broadly it can be seen as a different way of identifying the stance of an entity (person, organization, group). It can be applied whenever there is some aggregation of texts with regard to a set of entities. The distinction lies in the more fine-grained identification of stances: we (a) take larger chunks of text as input and (b) position the entities on a scale rather than characterizing them as in favor, neutral or against a given topic.

\section{Limitations}

The main limitation of the proposed study is the relatively small scale of the dataset it is based on. The proposed method is scalable and computationally undemanding (all of the analyzed models can be trained on a single GPU with
12G of memory), and it is feasible to apply it to other countries in the CMP dataset. However, in order to arrive at interpretable results that could be
verified in terms of policy substance based on the experts' knowledge of the political spectrum, we had to focus the evaluation part on the materials of a single election cycle in one country. Potentially, the method can be applied to any country whose manifestos have CMP annotations, however, further investigation with data from other countries needs to be carried out to verify that.

While most policies are recurrent in manifestos, there may be a few topics appearing in upcoming elections, adding some variability in debate across election years. The policy domain labeller might need to be updated every now and then with current topics of interest (e.g. Covid, a sudden expansion of the military). Therefore, the effect of news electoral programs in the classification step requires more investigation namely, the feasibility of further training with new topics of the current debate or the necessity to re-train the whole classifier with new manifestos over again. That being said, the CMP codebook has remained the same for over two decades now. We take this as evidence that the policy domains do not need to change, only the ability of the classifier to correctly identify sentences with unseen topics. 


%

\section*{Acknowledgements}
We are thankful for the insights on policy and party positioning contributed by Nils Düpont, Sebastian Haunss and Nico Blokker. We acknowledge funding by Deutsche Forschungsgemeinschaft (DFG) for project MARDY 2 (375875969) within the priority program RATIO.

\bibliography{anthology,custom}

\begin{thebibliography}{26}
\expandafter\ifx\csname natexlab\endcsname\relax\def\natexlab#1{#1}\fi

\bibitem[{Benoit and Laver(2006)}]{benoit2006party}
Kenneth Benoit and Michael Laver. 2006.
\newblock \emph{Party policy in modern democracies}.
\newblock Routledge.

\bibitem[{Bojanowski et~al.(2017)Bojanowski, Grave, Joulin, and
  Mikolov}]{bojanowski2017enriching}
Piotr Bojanowski, Edouard Grave, Armand Joulin, and Tomas Mikolov. 2017.
\newblock Enriching word vectors with subword information.
\newblock \emph{Transactions of the Association for Computational Linguistics},
  5:135--146.

\bibitem[{Braun and Schmitt(2020)}]{doi:10.1177/1354068818805248}
Daniela Braun and Hermann Schmitt. 2020.
\newblock \href {https://doi.org/10.1177/1354068818805248} {Different emphases,
  same positions? {T}he election manifestos of political parties in the {EU}
  multilevel electoral system compared}.
\newblock \emph{Party Politics}, 26(5):640--650.

\bibitem[{Budge(2003)}]{budge2003validating}
Ian Budge. 2003.
\newblock Validating the {Manifesto Research G}roup approach: theoretical
  assumptions and empirical confirmations.
\newblock In \emph{Estimating the policy position of political actors}, pages
  70--85. Routledge.

\bibitem[{Budge et~al.(2001)Budge, Klingemann, Volkens, Bara, and
  Tanenbaum}]{budge2001-mapping}
Ian Budge, Hans-Dieter Klingemann, Andrea Volkens, Judith Bara, and Eric
  Tanenbaum, editors. 2001.
\newblock \emph{Mapping {{Policy Preferences}}: {{Estimates}} for {{Parties}},
  {{Electors}}, and {{Governments}} 1945-1998}.
\newblock {Oxford University Press}, {Oxford, New York}.

\bibitem[{Burst et~al.(2021)Burst, Krause, Lehmann, Lewandowski, Matthieß,
  Merz, Regel, and Zehnter}]{manifesto-burst}
Tobias Burst, Werner Krause, Pola Lehmann, Jirka Lewandowski, Theres Matthieß,
  Nicolas Merz, Sven Regel, and Lisa Zehnter. 2021.
\newblock Manifesto corpus. version: 2021.1.
\newblock \emph{Berlin: WZB Berlin Social Science Center.}

\bibitem[{Ceron et~al.(2022)Ceron, Blokker, and Pad{\'o}}]{ceron-2022}
Tanise Ceron, Nico Blokker, and Sebastian Pad{\'o}. 2022.
\newblock \href {https://aclanthology.org/2022.conll-1.22} {Optimizing text
  representations to capture (dis)similarity between political parties}.
\newblock In \emph{Proceedings of the 26th Conference on Computational Natural
  Language Learning (CoNLL)}, pages 325--338, Abu Dhabi, United Arab Emirates
  (Hybrid). Association for Computational Linguistics.

\bibitem[{Conneau et~al.(2020)Conneau, Khandelwal, Goyal, Chaudhary, Wenzek,
  Guzm{\'a}n, Grave, Ott, Zettlemoyer, and
  Stoyanov}]{conneau-etal-2020-unsupervised}
Alexis Conneau, Kartikay Khandelwal, Naman Goyal, Vishrav Chaudhary, Guillaume
  Wenzek, Francisco Guzm{\'a}n, Edouard Grave, Myle Ott, Luke Zettlemoyer, and
  Veselin Stoyanov. 2020.
\newblock \href {https://doi.org/10.18653/v1/2020.acl-main.747} {Unsupervised
  cross-lingual representation learning at scale}.
\newblock In \emph{Proceedings of the 58th Annual Meeting of the Association
  for Computational Linguistics}, pages 8440--8451, Online. Association for
  Computational Linguistics.

\bibitem[{D{\"a}ubler and Benoit(2021)}]{daubler2021scaling}
Thomas D{\"a}ubler and Kenneth Benoit. 2021.
\newblock Scaling hand-coded political texts to learn more about left-right
  policy content.
\newblock \emph{Party Politics}, page 13540688211026076.

\bibitem[{Devlin et~al.(2019)Devlin, Chang, Lee, and
  Toutanova}]{devlin-etal-2019-bert}
Jacob Devlin, Ming-Wei Chang, Kenton Lee, and Kristina Toutanova. 2019.
\newblock \href {https://doi.org/10.18653/v1/N19-1423} {{BERT}: Pre-training of
  deep bidirectional transformers for language understanding}.
\newblock In \emph{Proceedings of the 2019 Conference of the North {A}merican
  Chapter of the Association for Computational Linguistics: Human Language
  Technologies, Volume 1 (Long and Short Papers)}, pages 4171--4186,
  Minneapolis, Minnesota. Association for Computational Linguistics.

\bibitem[{Gao et~al.(2021)Gao, Yao, and Chen}]{gao-etal-2021-simcse}
Tianyu Gao, Xingcheng Yao, and Danqi Chen. 2021.
\newblock \href {https://doi.org/10.18653/v1/2021.emnlp-main.552} {{S}im{CSE}:
  Simple contrastive learning of sentence embeddings}.
\newblock In \emph{Proceedings of the 2021 Conference on Empirical Methods in
  Natural Language Processing}, pages 6894--6910, Online and Punta Cana,
  Dominican Republic. Association for Computational Linguistics.

\bibitem[{Glava{\v{s}} et~al.(2017)Glava{\v{s}}, Nanni, and
  Ponzetto}]{glavas-etal-2017-unsupervised}
Goran Glava{\v{s}}, Federico Nanni, and Simone~Paolo Ponzetto. 2017.
\newblock \href {https://aclanthology.org/E17-2109} {Unsupervised cross-lingual
  scaling of political texts}.
\newblock In \emph{Proceedings of the 15th Conference of the {E}uropean Chapter
  of the Association for Computational Linguistics: Volume 2, Short Papers},
  pages 688--693, Valencia, Spain. Association for Computational Linguistics.

\bibitem[{Heywood(2021)}]{heywood2021political}
Andrew Heywood. 2021.
\newblock \emph{Political ideologies: An introduction}.
\newblock Bloomsbury Publishing.

\bibitem[{Iversen(1994)}]{10.2307/2111335}
Torben Iversen. 1994.
\newblock \href {http://www.jstor.org/stable/2111335} {Political leadership and
  representation in {West European} democracies: {A} test of three models of
  voting}.
\newblock \emph{American Journal of Political Science}, 38(1):45--74.

\bibitem[{Jolly et~al.(2022)Jolly, Bakker, Hooghe, Marks, Polk, Rovny,
  Steenbergen, and Vachudova}]{jolly2022chapel}
Seth Jolly, Ryan Bakker, Liesbet Hooghe, Gary Marks, Jonathan Polk, Jan Rovny,
  Marco Steenbergen, and Milada~Anna Vachudova. 2022.
\newblock {Chapel Hill} expert survey trend file, 1999--2019.
\newblock \emph{Electoral Studies}, 75:102420.

\bibitem[{Laver et~al.(2003)Laver, Benoit, and Garry}]{laver2003extracting}
Michael Laver, Kenneth Benoit, and John Garry. 2003.
\newblock Extracting policy positions from political texts using words as data.
\newblock \emph{American political science review}, 97(2):311--331.

\bibitem[{Laver and Budge(1992)}]{laver1992measuring}
Michael~J Laver and Ian Budge. 1992.
\newblock Measuring policy distances and modelling coalition formation.
\newblock In \emph{Party policy and government coalitions}, pages 15--40.
  Springer.

\bibitem[{Li et~al.(2020)Li, Zhou, He, Wang, Yang, and
  Li}]{li-etal-2020-sentence}
Bohan Li, Hao Zhou, Junxian He, Mingxuan Wang, Yiming Yang, and Lei Li. 2020.
\newblock \href {https://doi.org/10.18653/v1/2020.emnlp-main.733} {On the
  sentence embeddings from pre-trained language models}.
\newblock In \emph{Proceedings of the 2020 Conference on Empirical Methods in
  Natural Language Processing (EMNLP)}, pages 9119--9130, Online. Association
  for Computational Linguistics.

\bibitem[{Loshchilov and Hutter(2019)}]{loshchilov2019adamw}
Ilya Loshchilov and Frank Hutter. 2019.
\newblock \href {https://dblp.org/rec/conf/iclr/LoshchilovH19.html} {Decoupled
  weight decay regularization}.
\newblock In \emph{Proceedings of the 7th International Conference on Learning
  Representations, New Orleans, 6-9 May 2019}.

\bibitem[{Mantel(1967)}]{mantel1967detection}
Nathan Mantel. 1967.
\newblock The detection of disease clustering and a generalized regression
  approach.
\newblock \emph{Cancer research}, 27(2):209--220.

\bibitem[{Mikolov et~al.(2013)Mikolov, Chen, Corrado, and
  Dean}]{Mikolov2013EfficientEO}
Tomas Mikolov, Kai Chen, Gregory~S. Corrado, and Jeffrey Dean. 2013.
\newblock Efficient estimation of word representations in vector space.
\newblock In \emph{Proceedings of the International Conference on Learning
  Representations}.

\bibitem[{Reimers and Gurevych(2019)}]{reimers-gurevych-2019-sentence}
Nils Reimers and Iryna Gurevych. 2019.
\newblock \href {https://doi.org/10.18653/v1/D19-1410} {Sentence-{BERT}:
  Sentence embeddings using {S}iamese {BERT}-networks}.
\newblock In \emph{Proceedings of the 2019 Conference on Empirical Methods in
  Natural Language Processing and the 9th International Joint Conference on
  Natural Language Processing (EMNLP-IJCNLP)}, pages 3982--3992, Hong Kong,
  China. Association for Computational Linguistics.

\bibitem[{Rheault and Cochrane(2020)}]{rheault2020word}
Ludovic Rheault and Christopher Cochrane. 2020.
\newblock Word embeddings for the analysis of ideological placement in
  parliamentary corpora.
\newblock \emph{Political Analysis}, 28(1):112--133.

\bibitem[{Slapin and Proksch(2008)}]{slapin2008scaling}
Jonathan~B Slapin and Sven-Oliver Proksch. 2008.
\newblock A scaling model for estimating time-series party positions from
  texts.
\newblock \emph{American Journal of Political Science}, 52(3):705--722.

\bibitem[{Su et~al.(2021)Su, Cao, Liu, and Ou}]{Su2021WhiteningSR}
Jianlin Su, Jiarun Cao, Weijie Liu, and Yangyiwen Ou. 2021.
\newblock \href {https://arxiv.org/abs/2103.15316} {Whitening sentence
  representations for better semantics and faster retrieval}.
\newblock \emph{ArXiv}, abs/2103.15316.

\bibitem[{Waldherr et~al.(2019)Waldherr, Wehden, Stoltenberg, Miltner, Ostner,
  and Pfetsch}]{Waldherr_Wehden_Stoltenberg_Miltner_Ostner_Pfetsch_2019}
Annie Waldherr, Lars-Ole Wehden, Daniela Stoltenberg, Peter Miltner, Sophia
  Ostner, and Barbara Pfetsch. 2019.
\newblock \href {https://doi.org/10.17169/fqs-20.1.3058} {Inductive codebook
  development for content analysis: Combining automated and manual methods}.
\newblock \emph{Forum Qualitative Sozialforschung / Forum: Qualitative Social
  Research}, 20(1).

\end{thebibliography}
\bibliographystyle{acl_natbib}

\onecolumn
\appendix

\setcounter{table}{0}
\section{Data Statistics and Handling}
\label{sec:appendix-a}

\subsection{Data for the party positioning analysis}
\label{sec:appendix-positioning}

\begin{table}[H]
\centering
\small
\begin{tabular}{ll}
\textbf{Party}                                          & \textbf{2021} \\ \hline
The Left (Die Linke)                                    & 4850          \\
Social Democratic Party of Germany (SDP)                & 1665          \\
Alternative for Germany  (AfD)                          & 1574          \\
Christian Democratic Union/Christian Social Union (CDU) & 2775          \\
Alliance‘90/Greens (Grüne)                             & 3947          \\
Free Democratic Party (FDP)                             & 2239         
\end{tabular}
\caption{Number of sentences per party per year from the 2021 German elections. }
\label{tab:2021manifestos}
\end{table}

\subsection{Data for training the policy domain classifiers}
\label{sec:appendix-classifiers}

\paragraph{Preprocessing.} The CMP annotations contain the H and 0 labels for some sentences. While Hs are excluded from all the modelling because they represent the heading of a section. The 0 label is kept for the classifier in order to emulate a real world case scenario where there are labels that do not represent any policy domain/category. 

The \enquote{Germany} training regime with manifestos from Germany only contains 57,259 instances whereas the \enquote{German} regime with data from German-speaking countries has 106,724 instances in total. 10\% of each of them is used as the validation set.

\begin{table}[H]
\small
\centering
\begin{tabular}{llllll}
\textbf{}                                         & \textbf{2017} & \textbf{2002} & \textbf{2005} & \textbf{2009} & \textbf{2013} \\ \hline
Alliance‘90/Greens                                & 3826          & 1644          & 1860          & 3578          & 5382          \\
Alternative for Germany                           & 1003          & 0             & 0             & 0             & 72            \\
The Left                                          & 3926          & 0             & 0             & 1660          & 2453          \\
Free Democratic Party                             & 2053          & 1971          & 1398          & 2230          & 2560          \\
Party of Democratic Socialism                     & 0             & 840           & 0             & 0             & 0             \\
Christian Democratic Union/Christian Social Union & 1340          & 1293          & 769           & 1975          & 2534          \\
Social Democratic Party of Germany                & 2631          & 1591          & 880           & 2181          & 2873          \\
The Left. Party of Democratic Socialism           & 0             & 0             & 572           & 0             & 0             \\
Pirates                                           & 0             & 0             & 0             & 0             & 1755         
\end{tabular}
\caption{Number of sentences per party per year from the German elections. }
\label{tab:germany}
\end{table}

\begin{table}[H]
\small
\centering
\begin{tabular}{lllll}
\textbf{Party}                                        & \textbf{2007} & \textbf{2019} & \textbf{2011} & \textbf{2015} \\ \hline
Christian Democratic People’s Party of Switzerland & 125           & 313           & 148           & 278           \\
FDP. The Liberals                                   & 126           & 784           & 207           & 110           \\
Swiss People’s Party                               & 1035          & 1423          & 120           & 1329          \\
Conservative Democratic Party of Switzerland       & 0             & 974           & 72            & 329           \\
Swiss Labour Party                                 & 104           & 673           & 0             & 353           \\
Green Liberal Party                                & 94            & 144           & 68            & 225           \\
Christian Social Party                             & 172           & 0             & 270           & 0             \\
Social Democratic Party of Switzerland             & 1133          & 122           & 71            & 129           \\
Federal Democratic Union                           & 40            & 637           & 0             & 0             \\
Green Party of Switzerland                         & 800           & 571           & 411           & 506           \\
Protestant People’s Party of Switzerland           & 89            & 129           & 25            & 553          
\end{tabular}
\caption{Number of sentences per party per year from the Swiss elections. }
\label{tab:swiss}
\end{table}

\begin{table}[H]
\small
\centering
\begin{tabular}{lllllll}
\textbf{Party}                     & \textbf{2017} & \textbf{2019} & \textbf{2002} & \textbf{2006} & \textbf{2008} & \textbf{2013} \\ \hline
The New Austria and Liberal Forum  & 126           & 1170          & 0             & 0             & 0             & 1006          \\
Team Stronach for Austria          & 0             & 0             & 0             & 0             & 0             & 1195          \\
Austrian Communist Party           & 0             & 0             & 0             & 0             & 113           & 0             \\
Austrian People’s Party            & 2793          & 719           & 2163          & 2051          & 602           & 1157          \\
Austrian Freedom Party             & 452           & 220           & 2667          & 325           & 461           & 115           \\
Peter Pilz List                    & 71            & 0             & 0             & 0             & 0             & 0             \\
Austrian Social Democratic Party   & 2722          & 1893          & 1139          & 714           & 1189          & 716           \\
Alliance for the Future of Austria & 0             & 0             & 0             & 551           & 342           & 0             \\
The Greens                         & 1084          & 2248          & 683           & 693           & 691           & 2369         
\end{tabular}
\caption{Number of sentences per party per year from the Austrian elections. }
\label{tab:austria}
\end{table}

\begin{table}[H]
\tiny
\centering
\begin{tabular}{llllllll}
\textbf{Country} & \textbf{equality}                                                                      & \textbf{military and peace}                                                        & \textbf{\begin{tabular}[c]{@{}l@{}}democracy and \\ constitutionalism\end{tabular}} & \textbf{\begin{tabular}[c]{@{}l@{}}foreign relations, eu \\ and protectionism\end{tabular}} & \textbf{\begin{tabular}[c]{@{}l@{}}market regulation \\ and nationalisation\end{tabular}}                            & \textbf{\begin{tabular}[c]{@{}l@{}}political authority, \\ civic mindedness\\ and \\ anti-imperialism\end{tabular}} & \textbf{\begin{tabular}[c]{@{}l@{}}immigration, \\ multiculturalism \\ and human rights\end{tabular}} \\ \hline
Austria          & 3348                                                                                   & 555                                                                                & 2301                                                                                & 2369                                                                                        & 2181                                                                                                                 & 976                                                                                                                 & 1905                                                                                                  \\
Germany          & 5462                                                                                   & 1614                                                                               & 2784                                                                                & 3903                                                                                        & 5182                                                                                                                 & 2744                                                                                                                & 5094                                                                                                  \\
Switzerland      & 779                                                                                    & 403                                                                                & 431                                                                                 & 1388                                                                                        & 1218                                                                                                                 & 763                                                                                                                 & 1070                                                                                                  \\
Total            & 9589                                                                                   & 2572                                                                               & 5516                                                                                & 7660                                                                                        & 8581                                                                                                                 & 4483                                                                                                                & 8069                                                                                                  \\ \hline
\textbf{Country} & \textbf{\begin{tabular}[c]{@{}l@{}}labour groups \\ and \\ welfare state\end{tabular}} & \textbf{\begin{tabular}[c]{@{}l@{}}sustainability \\ and agriculture\end{tabular}} & \textbf{\begin{tabular}[c]{@{}l@{}}education \\ and \\ technology\end{tabular}}     & \textbf{\begin{tabular}[c]{@{}l@{}}culture and \\ civic mindedness\end{tabular}}            & \textbf{\begin{tabular}[c]{@{}l@{}}government admin, \\ (de)centralization \\ and economic \\ planning\end{tabular}} & \textbf{\begin{tabular}[c]{@{}l@{}}law and order \\ and traditional \\ morality\end{tabular}}                       & \textbf{other}                                                                                        \\ \hline
Austria          & 5222                                                                                   & 3288                                                                               & 4238                                                                                & 1476                                                                                        & 3450                                                                                                                 & 3131                                                                                                                & 224                                                                                                   \\
Germany          & 6386                                                                                   & 4311                                                                               & 5999                                                                                & 1484                                                                                        & 7865                                                                                                                 & 4022                                                                                                                & 409                                                                                                   \\
Switzerland      & 2022                                                                                   & 2198                                                                               & 1377                                                                                & 285                                                                                         & 1378                                                                                                                 & 1380                                                                                                                & 109                                                                                                   \\
Total            & 13630                                                                                  & 9797                                                                               & 11614                                                                               & 3245                                                                                        & 12693                                                                                                                & 8533                                                                                                                & 742                                                                                                  
\end{tabular}
\caption{Number of sentences per label and country for training the policy domain labeller.}
\label{tab:labels}
\end{table}

\subsection{Models' hyperparameters and libraries}

SBERT$_{frozen}$+Logistic Regression:
\begin{itemize}
    \item No hyperparameter optimization for the logistic regression model - default parameters from the library Scikit-learn
    \item Frozen SBERT model: \texttt{paraphrase-\-multilingual-mpnet-base-v2}
\end{itemize}
\bigskip

RoBERTa$_{xlm}$ + Multi-layer perception (MLP):
\begin{itemize}
    \item RoBERTa model: \texttt{xlm-roberta-large}
    \item First linear layer's input size: $\mathbb{R}^{Nx1024}$
    \item One tahn activation layer
    \item Second linear layer's input size: $\mathbb{R}^{Nx14}$
    \item 5 epochs
    \item AdamW optimizer \citep{loshchilov2019adamw}
    \item Learning rate: $10^{-5}$
    \item HuggingFace for implementation
\end{itemize}
\bigskip

SBERT$_{tune}$ + Multi-layer perception (MLP):
\begin{itemize}
    \item SBERT model: \texttt{paraphrase-\-multilingual-mpnet-base-v2}
    \item First linear layer's input size: $\mathbb{R}^{Nx768}$
        \item One tahn activation layer
    \item Second linear layer's input size: $\mathbb{R}^{Nx14}$
    \item 5 epochs
    \item AdamW optimizer \citep{loshchilov2019adamw}
    \item Learning rate: $10^{-5}$
    \item SBERT HuggingFace for implementation
\end{itemize}

Hardware information for all experiments:
\begin{itemize}
    \item System CPU: 2 x Intel Xeon E5-2650 v4, 2,20GHz, 12 Core
    \item 24 cores 
    \item 256 GByte of memory
    \item GPU: 4 x Nvidia GeForce GTX 1080 Ti, 12 GB
\end{itemize}

\section{Appendix}
\label{sec:appendix-b}

\subsection{Hierarchical clustering with CMP categories}

\begin{figure}[!htb]
    \centering
    \includegraphics[width=0.95\textwidth]{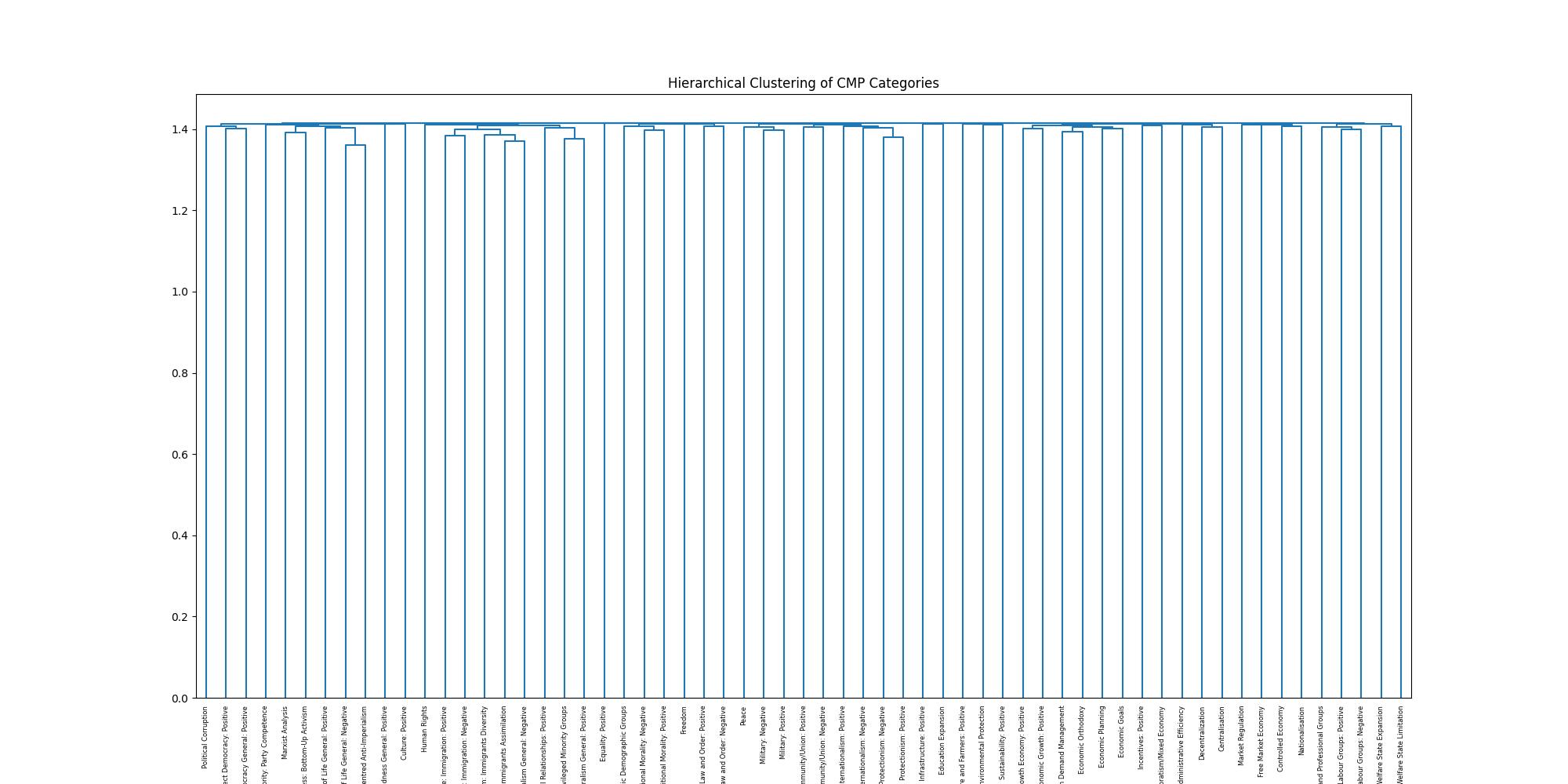}
    \caption{Results of the hierarchical clustering of lower-categories from the manifestos.}
    \label{fig:hirarchical-clustering}
\end{figure}

\newpage

\subsection{CMP categories clustered across Germany, Switzerland, and Austria}

\begin{table}[H]
\small
\centering
\begin{tabular}{ll}
\textbf{policy domain}                                                                                   & \textbf{Categories from CMP}                                                                                                                                                                                                                                                                                                                                                                                                                                                                                                                                                                                                                    \\ \hline
equality                                                                                                 & \textcolor{blue}{Equality: Positive}                                                                                                                                                                                                                                                                                                                                                                                                                                                                                                                                                                                                                              \\ \hline
military and peace                                                                                       & \textcolor{blue}{Military: Negative}, \textcolor{blue}{Peace}, \textcolor{blue}{Military: Positive}                                                                                                                                                                                                                                                                                                                                                                                                                                                                                                                                                                                                  \\ \hline
\begin{tabular}[c]{@{}l@{}}democracy and \\ constitutionalism\end{tabular}                               & \begin{tabular}[c]{@{}l@{}}\textcolor{blue}{Political Corruption}, \textcolor{blue}{Direct Democracy: Positive}, \textcolor{blue}{Democracy General: Positive}, \\ \textcolor{purple}{Constitutionalism: Negative}, \textcolor{purple}{Representative Democracy: Positive}, \\ \textcolor{purple}{Constitutionalism: Positive}, \textcolor{purple}{Democracy General: Negative}, Democracy\end{tabular}                                                                                                                                                                                                                                                                                                                                                                               \\ \hline
\begin{tabular}[c]{@{}l@{}}foreign relations, eu \\ and protectionism\end{tabular}                       & \begin{tabular}[c]{@{}l@{}}\textcolor{blue}{Internationalism: Negative}, \textcolor{blue}{European Community/Union: Positive}, \textcolor{blue}{Protectionism: Negative}, \\ \textcolor{blue}{Protectionism: Positive}, \textcolor{blue}{Internationalism: Positive}, \textcolor{blue}{European Community/Union: Negative}\end{tabular}                                                                                                                                                                                                                                                                                                                                                                                                                     \\ \hline
\begin{tabular}[c]{@{}l@{}}market regulation \\ and nationalisation\end{tabular}                         & \textcolor{blue}{Nationalisation, Controlled Economy, Free Market Economy, Market Regulation}                                                                                                                                                                                                                                                                                                                                                                                                                                                                                                                                                                     \\ \hline
\begin{tabular}[c]{@{}l@{}}political authority, \\ civic mindedness \\ and anti-imperialism\end{tabular} & \begin{tabular}[c]{@{}l@{}}\textcolor{blue}{Civic Mindedness: Bottom-Up Activism}, \textcolor{blue}{Political Authority: Party Competence}, \\ \textcolor{blue}{Anti-Imperialism: State Centred Anti-Imperialism}, \textcolor{blue}{Marxist Analysis}, \textcolor{blue}{National} \\ \textcolor{blue}{Way of Life General: Negative}, \textcolor{blue}{National Way of Life General: Positive}, \textcolor{purple}{Transition:} \\ \textcolor{purple}{Rehabilitation and Compensation}, \textcolor{purple}{Political Authority: Personal Competence}, \\ Political Authority, Political Authority: Strong government, Transition: Pre-Democratic \\ Elites: Negative, Civic Mindedness: Positive, Anti-Imperialism, Anti-Imperialism: \\ Foreign Financial Influence\end{tabular}                                                         \\ \hline
\begin{tabular}[c]{@{}l@{}}immigration, \\ multiculturalism and \\ human rights\end{tabular}             & \begin{tabular}[c]{@{}l@{}}\textcolor{blue}{National Way of Life: Immigration: Negative, Human Rights, Underprivileged Minority} \\ \textcolor{blue}{Groups, Multiculturalism General: Negative, Multiculturalism: Immigrants Assimilation}, \\ \textcolor{blue}{Foreign Special Relationships: Positive, Multiculturalism General: Positive}, \\ \textcolor{blue}{Multiculturalism: Immigrants Diversity, National Way of Life: Immigration: Positive}, \\ Freedom and Human Rights, \textcolor{purple}{Multiculturalism: Indigenous rights: Positive}, Multiculturalism: \\ Positive, National Way of Life: Positive, National Way of Life: Negative, Multiculturalism: \\ Negative, Foreign Special Relationships: Negative\end{tabular} \\ \hline
\begin{tabular}[c]{@{}l@{}}labour groups \\ and welfare state\end{tabular}                               & \begin{tabular}[c]{@{}l@{}}\textcolor{blue}{Welfare State Limitation, Middle Class and Professional Groups, Labour Groups: Positive}, \\ \textcolor{blue}{Labour Groups: Negative, Welfare State Expansion}\end{tabular}                                                                                                                                                                                                                                                                                                                                                                                                                                                            \\ \hline
\begin{tabular}[c]{@{}l@{}}sustainability \\ and agriculture\end{tabular}                                & \begin{tabular}[c]{@{}l@{}}\textcolor{blue}{Environmental Protection, Agriculture and Farmers: Positive, Sustainability: Positive}, \\ \textcolor{blue}{Agriculture and Farmers: Negative}, \textcolor{purple}{Agriculture and Farmers: Positive}\end{tabular}                                                                                                                                                                                                                                                                                                                                                                                                                                          \\ \hline
\begin{tabular}[c]{@{}l@{}}education and \\ technology\end{tabular}                                      & \textcolor{blue}{Technology and Infrastructure: Positive, Education Expansion}, Education Limitation                                                                                                                                                                                                                                                                                                                                                                                                                                                                                                                                                              \\ \hline
\begin{tabular}[c]{@{}l@{}}culture and \\ civic mindedness\end{tabular}                                  & \textcolor{blue}{Culture: Positive, Civic Mindedness General: Positive}                                                                                                                                                                                                                                                                                                                                                                                                                                                                                                                                                                                           \\ \hline
\begin{tabular}[c]{@{}l@{}}government admin, \\ (de)centralization \\ and economic planning\end{tabular} & \begin{tabular}[c]{@{}l@{}}\textcolor{blue}{Governmental and Administrative Efficiency, Corporatism/Mixed Economy}, \\ \textcolor{blue}{Anti-Growth Economy: Positive, Keynesian Demand Management, Centralisation}, \\ \textcolor{blue}{Economic Growth: Positive, Decentralization, Incentives: Positive, Economic Goals}, \\ \textcolor{blue}{Economic Planning, Economic Orthodoxy, Anti-Growth Economy: Positive}\end{tabular}                                                                                                                                                                                                                                                                                     \\ \hline
\begin{tabular}[c]{@{}l@{}}law and order and \\ traditional morality\end{tabular}                        & \begin{tabular}[c]{@{}l@{}}\textcolor{blue}{Law and Order: Negative, Traditional Morality: Negative, Non-economic} \\ \textcolor{blue}{Demographic Groups, Freedom, Law and Order: Positive, Traditional Morality:} \\ \textcolor{blue}{Positive}, Law and Order: Positive\end{tabular}                                                                                                                                                                                                                                                                                                                                                                                                              
\end{tabular}
\caption{Categories of CMP in final policy domain clusters. The ones in blue are the results of the policy domain grouping approach with SBERT whereas the ones in purple refer to the categories that occurred less than 10 times in the 2021 German manifestos, and therefore, are added manually in the clusters. The ones in black are also manually added because they were annotated in the manifestos used for the classification, but not for the analysis. }
\label{tab:clustering}
\end{table}

\end{document}